\pdfoutput=1

\documentclass[11pt]{article}

\usepackage[preprint]{coling}

\usepackage{times}
\usepackage{latexsym}
\usepackage{amsmath}
\usepackage{algorithm}
\usepackage{algorithmic}
\usepackage{booktabs}
\usepackage{multirow}
\usepackage{colortbl}
\usepackage[most]{tcolorbox}
\usepackage{xcolor,soul}
\usepackage{adjustbox}

\usepackage{xspace}
\usepackage[T1]{fontenc}

\usepackage[utf8]{inputenc}

\usepackage{microtype}

\usepackage{inconsolata}

\usepackage{graphicx}

\title{\ours: Structured Outputs Enable the Fast and Accurate \\Knowledge Editing for Large Language Models}

\author{Baolong Bi$^{1,2}$\quad Shenghua Liu$^{1,2,*}$\quad  Yiwei Wang$^3$\quad  Lingrui Mei$^{1,2}$ \\ \textbf{Hongcheng Gao}$^2$ \quad \textbf{Junfeng Fang}$^{4,*}$ \quad \textbf{Xueqi Cheng}$^{1,2}$ \\
	$^1$CAS Key Laboratory of AI Safety, Institute of Computing Technology, Chinese Academy of Sciences \\
        $^2$University of Chinese Academy of Sciences 
	$^3$University of California, Los Angeles \\
        $^4$University of Science and Technology of China (USTC), Hefei, China\\
        \small{\{bibaolong23z, liushenghua, cxq\}@ict.ac.cn},
        \small{wangyw.evan@gmail.com}	\\
        \small{\{meilingrui22, gaohongcheng23\}@mails.ucas.ac.cn, fjf@mail.ustc.edu.cn}
}

\newcommand{\dataa}{\textsc{MQuAKE}\xspace}
\newcommand{\datab}{\textsc{MQuAKE-3k}\xspace}
\newcommand{\datac}{\textsc{MQuAKE-2002}\xspace}
\newcommand{\datad}{\textsc{MQuAKE-hard}\xspace}

\newcommand{\llamaa}{\textsc{LLaMA2-7B-Chat}\xspace}
\newcommand{\llamab}{\textsc{LLaMA2-13B-Chat}\xspace}

\newcommand{\deep}{\textsc{DeepEdit}}
\newcommand{\gpta}{\textsc{GPT-3.5-Turbo-Instruct}\xspace}
\newcommand{\gptb}{\textsc{GPT-4o-mini}\xspace}
\newcommand{\ours}{{$\textsc{StruEdit}$}\xspace}

\definecolor{ggreen}{rgb}{0.0, 0.6, 0.0}
\definecolor{rred}{rgb}{0.75, 0.0, 0.0}
\definecolor{bblue}{rgb}{0.13, 0.67, 0.8}

\definecolor{darkred}{RGB}{200,0,0}
\definecolor{lightgreen}{RGB}{228,253,227}
\definecolor{lightred}{RGB}{252,231,234}
\definecolor{lightyellow}{RGB}{250,253,191}
\definecolor{lightblue}{RGB}{230,240,254}
\definecolor{lightorange}{RGB}{255,223,191}
\definecolor{white}{RGB}{255,255,255}

\begin{document}
\maketitle
\begin{abstract}

As the modern tool of choice for question answering, large language models (LLMs) are expected to deliver answers with up-to-date knowledge.
To achieve such ideal question-answering systems, locating and then editing outdated knowledge in the natural language outputs is a general target of popular knowledge editing methods.
However, this target is challenging, as both identifying which tokens to edit in the reasoning steps and ensuring the coherence of the revised reasoning chain are difficult tasks.
We argue that these challenges stem from the unstructured nature of natural language outputs.
To address the above challenges, we propose $\textbf{Stru}$ctural $\textbf{Edit}$ing (\ours), an improved baseline for knowledge editing. 
We first prompt LLMs to produce structured outputs consisting of reasoning triplets. Then, \ours removes any potentially outdated knowledge and efficiently refills the structured outputs with up-to-date information in a single step.
Experimental results show that \ours consistently delivers the highest accuracy with lowest latency compared with other knowledge editing methods.

\end{abstract}

\section{Introduction}

With the widespread deployment of large language models ~\citep[LLMs;][]{chatgpt, openai2023gpt4, DBLP:journals/corr/abs-2302-13971, touvron2023llama, song2024fmint}, their reliability in answering questions is crucial, which entails accurately responding to queries with up-to-date knowledge.
However, the knowledge used for pre-training LLMs cannot guarantee ongoing timeliness because the world is constantly changing.
Knowledge editing ~\citep[KE;][]{sinitsin2020editable, zhu2020modifying, de2021editing} has been proposed to update the knowledge for LLMs.  

\begin{figure}[t!]
  \includegraphics[width=\columnwidth]{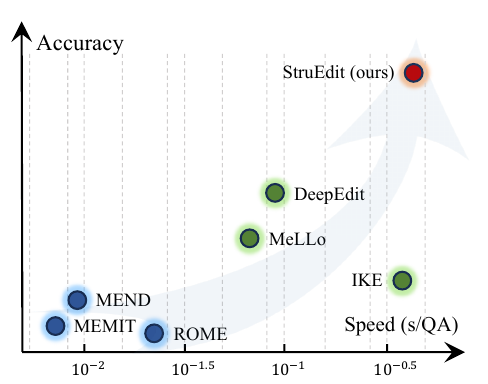}
  \vspace{-18pt}
  \caption{Comparison of performance between model editing (ME), in-context editing (ICE), and our \ours on multi-hop editing tasks, showing editing accuracy and average inference speed. Our \ours demonstrates the highest editing accuracy while maintaining the lowest latency.
  }
  \vspace{-16pt}
  \label{fig:compare}
\end{figure}

\begin{figure*}[t!]
  \includegraphics[width=\linewidth]{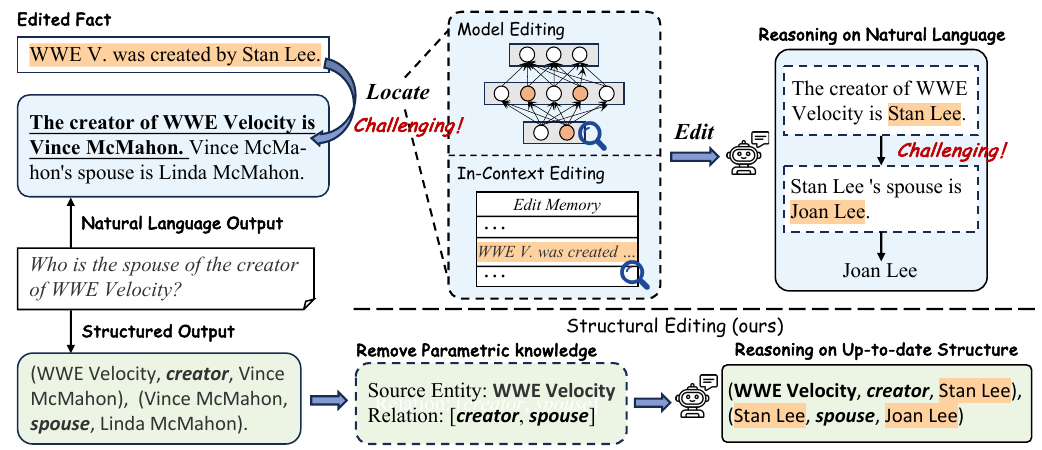}
  \caption{Differences between ME, ICE methods, and our structural editing. ME and ICE first locate the position of edited facts within the natural language reasoning steps (ME identifies modification regions, while ICE retrieves relevant new knowledge) before editing. Both face challenges with incorrect localization and inconsistent reasoning due to the natural language output format. In contrast, structural editing removes LLMs' parametric knowledge and reasons over up-to-date knowledge structures using structured output logic to derive the final answer.
  }
  \vspace{-10pt}
  \label{fig:overview}
\end{figure*}


The main process of existing KE methods can be summarized as \textit{Locate-Then-Edit}, which requires accurately reflecting specific edited facts within the natural language reasoning steps. In a chain-of-thought (CoT)~\citep{zhang2022automatic} process, this means adjusting certain natural language reasoning steps based on new knowledge and accurately inferring the final result using that updated information.
Model editing ~\citep[ME;][]{meng2022locating, meng2022mass, mitchell2022memory, yao2023editing, xu2024editing} locates the position of knowledge to be edited, such as neurons in the FFN or matrix regions, and modifies them.
In-context editing \citep[ICE;][]{madaan2022memory, zhong2023mquake, zheng2023can, wang2024deepedit, bi2024decoding, bi2024adaptive} locates relevant passages in the edit memory, prompting LLMs to utilize new knowledge to answer questions.

However, it is difficult to identify the tokens that need editing within the natural language reasoning steps, and incorrectly modifying parameters or providing inaccurate knowledge can directly result in editing failure.
Additionally, editing the tokens while ensuring the coherence of the output reasoning chain is challenging, as conflicts between new knowledge and parametric knowledge \citep{petroni2020contextaffectslanguagemodels, si2023promptinggpt3reliable, xie2024adaptivechameleonstubbornsloth} can lead to hallucinations during the reasoning process \citep{zhang2023sirens, huang2023survey, wang2023survey} or make stubborn knowledge difficult to edit \citep{bi2024decoding}.

In this paper, we argue that existing KE methods pose risks due to the \textit{Locate-Then-Edit} approach based on natural language reasoning. 
We propose a new paradigm, structural editing, which structures the natural language outputs. 
Instead of relying on the two-step process of locating and editing, we directly remove all information potentially affected by new knowledge and refill the output based on the updated information. 
This approach eliminates the challenges caused by the coupling of different reasoning steps, enabling multi-step edits to be completed in a one-shot manner.
Figure \ref{fig:overview} shows the differences between structural editing and previous methods.
To assess these approaches, we observe their performance on multi-hop editing tasks.
We found that ME and ICE methods perform poorly when batch\_size=full, with accuracy dropping significantly as the number of hops increases, indicating their difficulty in thoroughly editing knowledge.
In contrast, the new structural editing demonstrates a high success rate and robustness, showcasing its potential.


Building on these insights, we propose an effective improved baseline for knowledge editing, called \ours. 
\ours edits LLM outputs through knowledge structures without the need to locate outdated knowledge and also the corresponding model parameters or input context.
We use LLMs to refill new knowledge into the triplet reasoning structure based on specific logical rules, which accelerates reasoning speed and eliminates issues like hallucinations.
Specifically, we extract the source entity and sequential relations from the reasoning chain, perform entity matching, and select relations in the knowledge structure to infer the reasoning path and obtain the final answer.

\begin{table*}[t!]
\centering
\renewcommand{\arraystretch}{1.3}
\resizebox{\linewidth}{!}{
\begin{tabular}{llcccccccc}
\toprule
\multirow{2}{*}{\textbf{Model}} & \multirow{2}{*}{\textbf{Method}} & \multicolumn{4}{c}{\textbf{batch\_size=1}} & \multicolumn{4}{c}{\textbf{batch\_size=full}} \\ 
\cmidrule(lr){3-6} \cmidrule(lr){7-10}
 &  & \textit{2-hop} & \textit{3-hop} & \textit{4-hop} & \textit{avg.} & \textit{2-hop} & \textit{3-hop} & \textit{4-hop} & \textit{avg.}\\ 
\midrule
 & ROME$^\spadesuit$~\citep{meng2022locating} & 35.4 & 20.3 & 16.2 & 23.9 & 4.2 & 2.5 & 0.7 & 2.5 \\ 
 \multirow{2}{*}{\textsc{LLaMA2-}} & MEMIT$^\spadesuit$~\citep{meng2022mass} & 27.3 & 13.5 & 8.2 & 16.3 & 5.7 & 2.8 & 1.1 & 3.2  \\ 
 \cmidrule(lr){2-10}
 \multirow{2}{*}{\textsc{7B-chat}} & IKE$^\diamondsuit$~\citep{zheng2023can} & 80.8 & 63.8 & 50.9 & 65.2 & 13.5 & 5.7 & 2.6 & 7.3 \\ 
 & MeLLo$^\diamondsuit$~\citep{zhong2023mquake} & 54.9 & 34.7 & 30.2 & 39.9 & 29.9 & 9.2 & 3.1 & 14.1 \\ 
 \cmidrule(lr){2-10}
 & Structural Editing (ours) & \textbf{100} & \textbf{100} & \textbf{100} & \textbf{100} & \textbf{91.5} & \textbf{90.7} & \textbf{56.8} & \textbf{79.1} \\ 
\midrule
 \multirow{2}{*}{\textsc{GPT-3.5-turbo}} & IKE$^\diamondsuit$~\citep{zheng2023can} & 78.5 & 76.2 & 73.4 & 76.0 & 17.3 & 9.6 & 6.7 & 11.2\\ 
 \multirow{2}{*}{\textsc{-instruct}} & MeLLo$^\diamondsuit$~\citep{zhong2023mquake} & 72.6 & 48.7 & 40.5 & 53.9 & 47.8 & 20.2 & 16.8 & 28.3\\ 
 \cmidrule(lr){2-10}
 & Structural Editing (ours) & \textbf{100} & \textbf{100} & \textbf{100} & \textbf{100} & \textbf{98.9} & \textbf{97.7} & \textbf{95.8} & \textbf{97.4}\\ 
\bottomrule
\end{tabular}
}
\caption{Experimental results (accuracy; \%) on \datac for multi-hop editing tasks (2, 3, 4-hop). We evaluated both open-source and closed-source LLMs across ME, ICE methods, and our \ours. Methods marked with $^\spadesuit$ belong to ME, while those marked with $^\diamondsuit$ belong to ICE. The best editing result on every LLM is highlighted in \textbf{bold} font.}
\label{tab:observation}
\end{table*}

Experimental results demonstrate that our \ours consistently achieves the highest editing accuracy and the fastest speed compared to existing KE methods, as shown in Figure \ref{fig:compare}.  
\ours maintains robust editing capabilities as the number of reasoning hops and edited instances increases.
Our work provides an improved KE baseline for LLMs with higher accuracy, faster performance, and greater robustness, paving the way for further advancements in KE.



\section{Knowledge Editing on Multi-Hop Editing Tasks}
\label{sec:ob}

In this paper, we focus on multi-hop editing tasks. 
Single-hop fact editing, such as modifying a fact triplet $(s, r, t)$ to $(s, r, t')$, has been effectively addressed~\citep{wang2023easyedit}. 
However, in real-world knowledge question answering (QA), changing one fact can trigger a “ripple effect” requiring updates to additional related facts~\citep{cohen2024evaluating}. 
Therefore, in multi-hop editing tasks, it is crucial for LLMs to accurately reason the correct answer without introducing hallucinations caused by conflicts with parametric knowledge.

\subsection{Multi-Hop Editing}

Multi-hop editing is a more challenging task in KE, where LLMs need to consistently account for both the edited facts and related fact updates during multi-hop reasoning to ensure thorough knowledge revision. 
The main challenge lies in the potential conflict between new knowledge and the parametric knowledge in LLMs, which can result in factual hallucinations during reasoning~\citep{bi2024factuality}.
For instance, regarding a two-hop fact chain (\textit{WWE Velocity}, \textit{created by}, \textit{Vince McMahon}), (\textit{Vince McMahon}, \textit{spouse}, \textit{Linda McMahon}).
With a fact edit $(\textit{WWE Velocity}, \textit{created by}, \textit{Stan Lee})$ and an additional fact chain $(\textit{Stan Lee}, \textit{spouse}, \textit{Joan Lee})$, the correct updated answer should be \textit{Joan Lee}.

\subsection{Evaluation and Analysis}
\label{sec:ob_eva}

To thoroughly explore the editing capabilities of the main KE methods, including ME, ICE, and the new structural editing paradigm proposed in this paper, we conduct multi-hop editing experiments on the \dataa dataset (see Section \ref{sec:dataset} for details) with both open-source (\llamaa) and closed-source (\gpta) models.
Specifically, we construct multi-hop fact chains from the dataset and edit them with new knowledge based on each method.
We set the batch size of the edit memory to 1 and full batch for KE evaluation.
The batch size refers to the number of instances providing the edited facts for knowledge retrieval.  
A batch size of 1 means only the new knowledge relevant to the reasoning is provided, while a full batch simulates a real-world editing scenario where all new knowledge is provided, even if it is not directly related to the current reasoning task.

Table \ref{tab:observation} presents the results of this experiment. 
From our observations, we found that structural editing consistently achieves the best performance. 
Notably, in the full batch size knowledge memory setting, ME and ICE methods perform poorly, while structural editing shows a significant lead on both open-source and closed-source models.

\begin{figure*}[t!]
  \includegraphics[width=\linewidth]{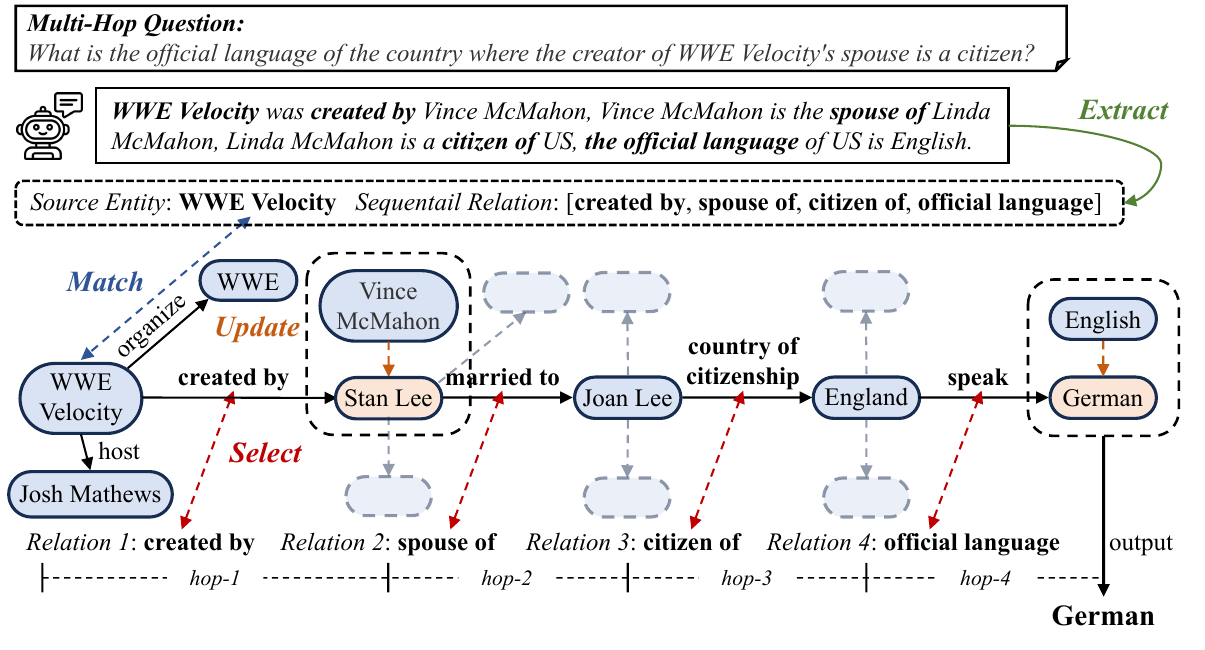}
  \vspace{-20pt}
  \caption{An illustration showing how \ours answers multi-hop questions using new knowledge. For a multi-hop question, \ours first guides LLMs to generate a reasoning chain using their parametric knowledge. It then extracts the source entity and sequential relations, matches the source entity within an external knowledge structure, and selects based on the sequential relations during reasoning to arrive at the final answer.
  }
  \vspace{-2mm}
  \label{fig:framework}
\end{figure*}

Furthermore, structural editing shows robust performance across varying batch sizes and reasoning hops compared to other methods.
First, all methods show a noticeable drop in average accuracy when moving from batch\_size = 1 to batch\_size = full. 
In the ICE methods, IKE experiences the largest drop due to its struggle in retrieving effective new knowledge for complex reasoning chains, while MeLLo is less affected as it breaks down multi-hop queries into sequential single-hop queries. 
In the ME methods, additional parameter edits can lead to hallucinations. 
Our proposed structural editing method shows the smallest drop because, unlike other methods, it does not require locating specific knowledge, making it less affected by the number of editing instances.

As the number of reasoning hops increases, the accuracy of both ME and ICE methods decreases to varying extents. 
Structural editing maintains 100\% accuracy when batch\_size=1 because, with the given structured fact chains and edited facts, LLMs can easily perform single-chain reasoning. 
Even with a full batch size, it remains stable, despite the need to consider more possible reasoning paths. 
This stability is due to the standardized knowledge representation in structured formats, which, compared to text, provides more reliable knowledge for LLMs' reasoning and reduces factual hallucinations.

\section{\ours: An Improved Baseline of Knowledge Editing}

Section \ref{sec:ob} demonstrates the exceptional accuracy and robustness of structural editing in multi-hop editing tasks. 
To address more generalized multi-hop QA problems with new knowledge, we propose a more comprehensive method, \ours, an improved baseline for knowledge editing. 
The main idea behind ME and ICE methods is to combine edited facts with parametric knowledge, relying on the strong reasoning capabilities of LLMs to answer questions. 
However, this approach is implicit, as it’s unclear whether the model’s parameters were correctly updated or if the new knowledge is trusted by the LLMs. 
To reduce the burden on LLMs and the uncertainty of editing, \ours does not retain parametric knowledge and no longer targets specific knowledge for editing.
Instead, all related knowledge is updated, allowing LLMs to reason over up-to-date knowledge structures based on the extracted logic of the question. Figure \ref{fig:framework} illustrates the framework of \ours, using KGs as an example of the knowledge structure.
We introduce the details of \ours from the following aspects.




\subsection{Structrual Editing on Parametric Output}

\ours uses up-to-date knowledge from the knowledge structure to edit LLMs' parametric output for multi-hop QA, leveraging the  logic rules to reason over the structure.
To enable reasoning over the knowledge structure for multi-hop questions, it is essential to provide the necessary conditions, including the source entity and sequential relations.
First, we input the initial multi-hop question into the LLMs and use in-context demonstrations to guide them in generating a multi-hop reasoning chain. 
Then, we extract the source entity and sequential relations from this chain using LLMs, providing logic for subsequent reasoning.

In this process, we discard all other entities in the chain reflecting parametric knowledge without checking for conflicts with the new knowledge. 
We only utilize LLMs to obtain invariant relations, which are invariant over time in the reasoning chain, and entities are very unstable as connections therein.
This reflects the core of our \ours, where we directly remove all information potentially affected by new knowledge and then refill it based on the updated knowledge. 
This ensures efficient and explicit editing.

\subsection{Multi-hop Reasoning with LLMs}

Multi-hop reasoning over a knowledge structure is key to our \ours approach.
Formally, given a source entity $e_0^t$ and a sequential relation $R=\{r_1^t, ..., r_h^t\}$ extracted according to \ref{sec:prep} for an $h$-hop question, we ideally aim to find a reasoning path $P^t=\{(e_0^t, r_1^t, e_1), ..., (e_{h-1}, r_h^t, e_h)\}$ that leads to the final answer $e_h$.

However, although the source entity and sequential relations provide the reasoning logic, the accuracy of reasoning can be significantly impacted by discrepancies between the entities and relations extracted from the reasoning chain and those in the knowledge structure.
For instance, as shown in Figure \ref{fig:framework}, "spouse of" in the sequential relation does not align with "married to" in the knowledge structure, which could lead to the selection of an alternate path during reasoning, ultimately resulting in an incorrect outcome.
To address this issue, and inspired by \citet{bi-etal-2024-lpnl}, we adopt the following strategies when entities and relations cannot be precisely matched.

\begin{figure}[t!]
    \centering
\begin{tcolorbox}[fonttitle=\bfseries, title=Entity / Relation Query Template, size = normal, label=mybox, center title]
\textbf{prefix\_question}: Which candidate entity/relation best matches the entity $e_0$ / relation $r_i^t$ <\textit{feature}>?   
\tcbline

\textbf{candidate\_description}: c$_{1}$: <\textit{feature}>, c$_{2}$: <\textit{feature}>, ..., 
c$_{|C|}$: <\textit{feature}>
\end{tcolorbox}
\caption{The query template has two components: prefix\_question, a selective question, and candidate\_description, describing the candidate set $C={c_1, c_2, ..., c_{|C|}}$, which represents either all entities or the relations associated with $e_{i-1}$. <\textit{feature}> denotes the textual description of entities or relations.}
\label{fig:template}
\vspace{-12pt}
\end{figure}

\paragraph{Entity Matching}

We construct a candidate set containing all entities, then query the LLMs with $e_0^t$ to identify the most closely matching entity.

\paragraph{Relation Selection}

Similarly, during the $i$-hop reasoning, we construct a candidate set based on all relations $\{r_i^1, ..., r_i^m\}$ associated with the entity ${e_{i-1}}$ to select the relation most similar to $r_i^t$.

The query template for entity matching and relation selection is shown in Figure \ref{fig:template}.
Entities and relations are aligned through LLM queries to optimally infer a reasoning path $P=\{(e_0, r_1, e_1), ..., (e_{h-1}, r_h, e_h)\}$ within the knowledge structure, leading to the final answer $e_h$, where $e_0$ best matches the extracted $e_0^t$ and $\{r_1, ..., r_h\}$ most closely correspond to the extracted $\{r_1^t, ..., r_h^t\}$.

\section{Experiments}
\subsection{Datasets and Tasks}
Unlike the evaluation editing tasks in Section \ref{sec:ob_eva}, we assess KE performance in the form of more generalized question answering tasks.
We focus exclusively on the more realistic and challenging multi-hop tasks to assess whether the knowledge has been thoroughly edited.
We conduct experiments using \datab \citep{zhong2023mquake} along with its challenging derivatives, \datac and \datad, introduced by \citet{wang2024deepedit}. 
\dataa is a multi-hop QA benchmark for knowledge editing that provides multi-hop knowledge questions to evaluate KE on counterfactual edits.
We construct KGs from the knowledge triples provided in \dataa to serve as the knowledge structure for our \ours.

\label{sec:dataset}

\begin{table*}[t]
\centering
\begin{adjustbox}{width=\linewidth}
\setlength{\tabcolsep}{8pt}
\renewcommand{\arraystretch}{0.9}
\begin{tabular}{@{}llcccc@{}}
\toprule
\textbf{Model} & {\textbf{Method}} &  {\textbf{\textsc{MQuAKE-3k}}} &{\textbf{\textsc{MQuAKE-2002}}} & {\textbf{\textsc{MQuAKE-hard}}}\\

\midrule
 & ROME$^\spadesuit$ \cite{mitchell2021fast}  & 2.3 &  2.9 & 0.4\\
 \multirow{2}{*}{\textsc{LLaMA2-}} & MEMIT$^\spadesuit$ \cite{meng2022mass}  & 3.1 & 3.5  & 0.6\\
 & MEND$^\spadesuit$ \cite{meng2022locating}
 & 3.9 &  4.1 & 0.9\\
  \cmidrule(lr){2-5}
 \multirow{2}{*}{\textsc{7B-chat}} & IKE$^\diamondsuit$ \cite{zheng2023can}  & 6.2 & 6.5  & 0.5\\
 & MeLLo$^\diamondsuit$ \cite{zhong2023mquake}  & 10.8 & 11.8  & 1.6\\
 &\deep$^\diamondsuit$ \cite{wang2024deepedit} & 11.2 & 12.9 & 7.0\\
 \cmidrule(lr){2-5}
 &\ours (ours) & \textbf{52.1} & \textbf{67.3} & \textbf{41.7}\\
\midrule
 & ROME$^\spadesuit$ \cite{mitchell2021fast}  & 3.1 &  4.8 & 0.7\\
 \multirow{2}{*}{\textsc{LLaMA2-}} & MEMIT$^\spadesuit$ \cite{meng2022mass}  & 4.3 & 5.1  & 1.1\\
 & MEND$^\spadesuit$ \cite{meng2022locating} & 4.8 &  5.3 & 1.3\\
 \cmidrule(lr){2-5}
 \multirow{2}{*}{\textsc{13B-chat}} & IKE$^\diamondsuit$ \cite{zheng2023can}  & 6.8 & 7.7  & 1.2\\
 & MeLLo$^\diamondsuit$ \cite{zhong2023mquake}  & 11.2 & 12.3  & 1.5\\
 &\deep$^\diamondsuit$ \cite{wang2024deepedit} & 12.5 & 13.7 & 8.2\\
 \cmidrule(lr){2-5}
 &\ours (ours) & \textbf{53.4} & \textbf{68.5} & \textbf{48.9}\\
\bottomrule
		\end{tabular}
	\end{adjustbox}
	\caption{Experimental results (accuracy; \%) on \dataa datasets with open-source models.
    We conduct the experiments with the full batch size edit memory.
    Methods marked with $^\spadesuit$ belong to ME, while those marked with $^\diamondsuit$ belong to ICE.
    The best KE result on every LLM is highlighted in \textbf{bold} font. 
    \label{tab:open}}
\end{table*}

\subsection{Models and Baselines}
We examine both closed-source models, including \llamaa and \llamab, as well as open-source models, including \gpta and \gptb.
We use state-of-the-art ME and ICE methods as our baselines for comparison with our \ours, which include the following approaches:
\paragraph{ROME} ROME~\citep{meng2022locating} applies causal mediation analysis to locate the editing area, framing model editing as a least-squares problem under linear equality constraints and solving it using lagrange multipliers.
\vspace{-2pt}
\paragraph{MEND} MEND~\citep{mitchell2021fast} adopt a meta-learning approach that trains a hypernetwork to infer weight updates from the gradient of the inserted fact.
\vspace{-2pt}
\paragraph{MEMIT} MEMIT ~\citep{meng2022mass} insert new memories into language models by targeting key transformer weights identified as causal mediators of factual knowledge recall.
\vspace{-2pt}
\paragraph{IKE} IKE ~\citep{zheng2023can} uses demonstration contexts without parameter updates, prompting LLMs to perform edits by leveraging newly retrieved knowledge.
\vspace{-2pt}
\paragraph{MeLLo} MeLLo ~\citep{zhong2023mquake} guides LLMs in multi-hop knowledge editing by decomposing subproblems and detecting conflicts between parametric knowledge and edited facts.
\vspace{-2pt}
\paragraph{\deep} \deep ~\citep{wang2024deepedit} enhances generating coherent reasoning chains with new knowledge through depth-first search.

\subsection{Overall Performance}

We set the edit memory to full batch size to reflect real-world scenarios in our experiments.
Table \ref{tab:open} displays the KE performance of ME and ICE methods, as well as our \ours, on \textsc{MQuAKE} across open-source models.
Overall, both ME and ICE methods perform poorly, while our \ours consistently shows a significant lead.
ME methods rely on modifying model structures or parameters to update knowledge, which becomes inefficient when dealing with a large number of new knowledge instances, as in our full batch size experiment. 
This can negatively impact the model's inherent parametric knowledge and reasoning abilities. 
In the ICE methods, IKE struggles to retrieve relevant information from the vast amount of new knowledge, resulting in poor editing performance. 
Although MeLLo and \deep attempt to address this limitation through conflict detection and deep search, they are still constrained by the reasoning capabilities of open-source LLMs.
Our \ours demonstrates significant improvement, highlighting its great potential in real-world scenarios.

\begin{table*}[t!]
\centering
\small
\renewcommand{\arraystretch}{1.2}
\resizebox{\linewidth}{!}{
\begin{tabular}{llccc}
\toprule
\textbf{Model} & {\textbf{Method}} &  {\textbf{\textsc{MQuAKE-3k}}} &{\textbf{\textsc{MQuAKE-2002}}} & {\textbf{\textsc{MQuAKE-hard}}}\\
\midrule
& IKE$^\diamondsuit$~\citep{zheng2023can} & 10.2 & 12.5 & 1.4 \\
 {\textsc{GPT-3.5}}& MeLLo$^\diamondsuit$ \cite{zhong2023mquake} & 20.0 & 25.1 & 1.6 \\
{\textsc{Turbo-Instruct}} &\deep$^\diamondsuit$ \cite{wang2024deepedit} & 38.0 & 48.0 & 53.7 \\
&\ours (ours) & \bf 62.7 & \bf 85.3 & \bf 75.5 \\
\midrule
& IKE$^\diamondsuit$~\citep{zheng2023can} & 10.5 & 13.1 & 1.5 \\
 \multirow{2}{*}{\textsc{GPT-4o-mini}}& MeLLo$^\diamondsuit$ \cite{zhong2023mquake} & 21.2 & 27.8 & 2.4 \\ &\deep$^\diamondsuit$ \cite{wang2024deepedit} & 41.3 & 49.2 & 55.8 \\
&\ours (ours) & \bf 66.5 & \bf 86.3 & \bf 77.3 \\
\bottomrule
\end{tabular}
}
\caption{Experimental results (accuracy; \%) of ICE methods and our \ours on \dataa datasets with closed-source models. We conduct the experiments with the full batch size edit memory.}
\label{tab:close}
\end{table*}

The experimental results on the closed-source models are presented in Table \ref{tab:close}.
Due to the closed-source nature of the models, we are only able to test the ICE methods and our \ours. 
The ICE methods exhibit overall improvement compared to their KE performance on open-source models, as  shown in Table \ref{tab:open}, with \deep showing the most significant gains. 
This improvement is attributed to the strong in-context learning (ICL) capabilities of the advanced closed-source models. 
However, our \ours also benefits from the enhanced ICL abilities of these LLMs, consistently achieving the best performance with an average accuracy that is 58.8\% higher than \deep.

Compared to more powerful closed-source models, \ours shows an even greater lead on smaller open-source models.  
This indicates that previous KE methods heavily relied on the inherent capabilities of LLMs, while \ours reduces the burden on LLMs during the KE process.

\subsection{Robustness of Knowledge Editing}

The robustness of KE is crucial for assessing whether knowledge has been thoroughly and effectively edited.
We evaluate the robustness of knowledge editing by assessing it across both the number of hops in multi-hop QA and the number of edited instances.

\paragraph{Number of Hops in Multi-Hop QA}

\begin{figure}[t]
    \centering
    \includegraphics[width=\linewidth]{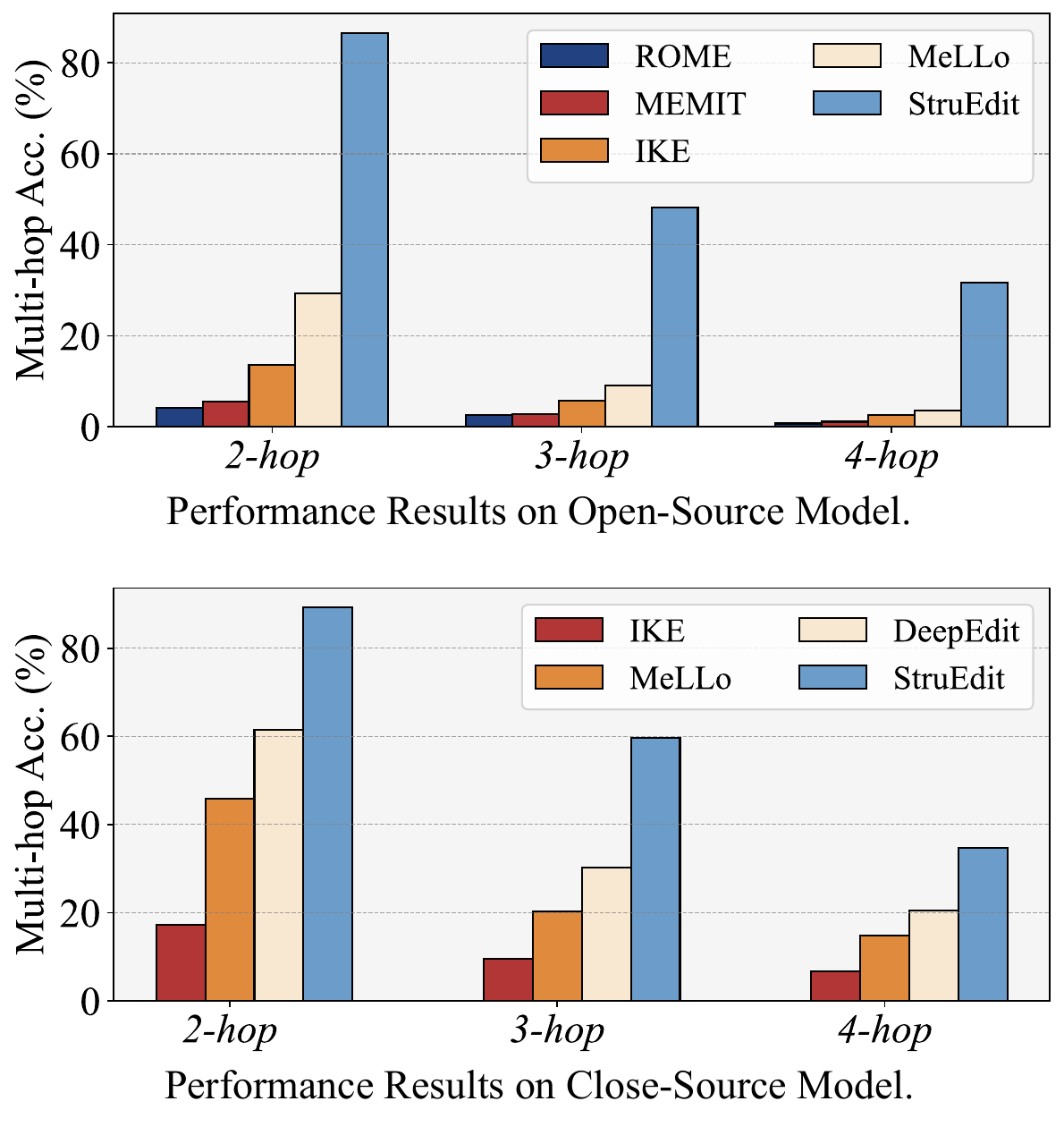}
    \vspace{-6mm}
    \caption{Multi-hop QA results across 2, 3, and 4 hops on both open-source (\llamaa) and closed-source (\gpta) models for ME, ICE methods, and our \ours.}
    \vspace{-6mm}
    \label{fig:hopnum}
\end{figure}

Figure \ref{fig:hopnum} shows the changes in KE performance for ME, ICE methods, and our \ours as the number of hops in multi-hop QA increases. 
We observed that, regardless of whether on open-source or closed-source models, all methods experience a noticeable decline in editing performance as the number of hops increases. 
This decline is primarily due to the hallucinations introduced by multi-hop reasoning and knowledge conflicts. 
Specifically, ME shows an average decrease of 57\% and ICE an average decrease of 56\% with each additional hop, whereas \ours only declines by 38\% on average, demonstrating the strong robustness of our \ours with increasing reasoning hops.

\paragraph{Number of Edited Instances}

\begin{figure*}[t!]
    \centering
    \includegraphics[width=\linewidth]{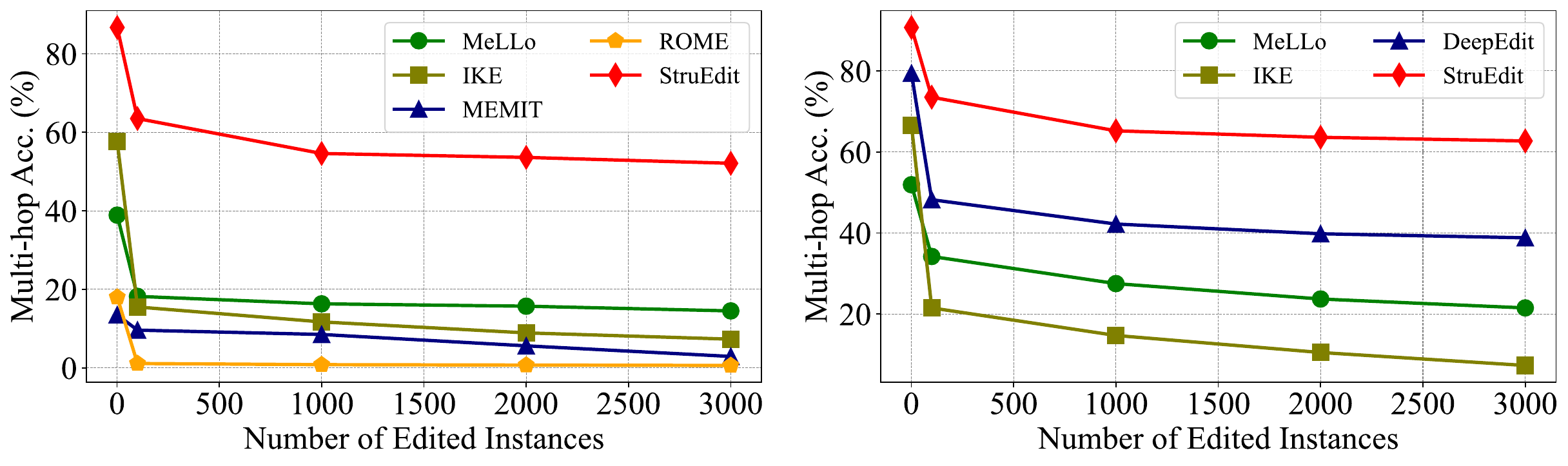}
    \vspace{-6mm}
    \caption{Multi-hop performance (CoT) of \llamaa (left) and \gpta (right) across different KE methods with 1, 100, 1000, 2000, and 3000 edited instances drawn for editing}
    \label{fig:batch_size}
\end{figure*}

Edited instances refer to the number of new knowledge updates required in the edit memory.
In real-world deployments, where the number of edited instances is often high, the robustness of KE in this aspect becomes especially critical.
We conduct experiments based on randomly grouped edited instances of varying quantities, with the results shown in Figure \ref{fig:batch_size}.
Consistent with the observations of \citet{zhong2023mquake}
, all methods show further decline when more edits are injected.
As the number of edited instances increases, both ME and ICE methods experience a significant decline, particularly in the comparison between 1-instance and 100-instance settings. 
This decline is primarily due to the challenges of imprecise knowledge localization, which fails to provide effective editing information, and hallucinations caused by knowledge conflicts. 
In contrast, our \ours consistently demonstrates the best performance across different instance scenarios, with the smallest average decline.

Overall, the performance of existing KE methods is significantly impacted by more complex multi-hop reasoning and a higher number of edited instances, which implies that many pieces of knowledge may not be thoroughly edited in real-world scenarios.
In contrast, our \ours demonstrates more stable robustness by reasoning over the updated knowledge structure, effectively mitigating hallucinations caused by retrieval errors, reasoning challenges, and knowledge conflicts.

\subsection{Editing Latency}
\label{sec:latency}

Table \ref{tab:latency} shows the latency of different editing methods in multi-hop QA.  
ME methods are generally slower, as it requires locating and modifying the model based on the edited facts before reasoning.  
ICE methods are faster, but latency increases when longer text reasoning is needed to improve editing performance, as seen in methods like MeLLo and \deep.  
\ours demonstrates the best efficiency, even compared to the simplest IKE method, because it achieves strong reasoning performance without relying on LLMs generating lengthy CoT, thanks to the support of knowledge structures.

\begin{table}[h]
\centering
\small
\scriptsize
\renewcommand{\arraystretch}{0.3}
\resizebox{\linewidth}{!}{
\begin{tabular}{lccc}
\toprule
\textbf{Method} & \textit{2-hop} & \textit{3-hop} & \textit{4-hop} \\
\midrule
MEND$^\spadesuit$ & 91.7 & 126.3 & 167.2\\
 ROME$^\spadesuit$ & 43.7 & 52.3 & 63.8\\
 MEMIT$^\spadesuit$ & 113.6 & 155.2 & 213.4\\
 \midrule
 IKE$^\diamondsuit$ & 2.48 & 2.67 & 2.85 \\
 MeLLo$^\diamondsuit$ & 12.83 & 16.27 & 21.53 \\
 \deep$^\diamondsuit$ & 10.05 & 15.79 & 19.58 \\
 \midrule
 \ours & \bf 1.75 & \bf 2.25 & \bf 2.38 \\
\bottomrule
\end{tabular}
}
\caption{Average latency (s/QA) for KE on \datac with \llamaa. The lowest latency is highlighted in \textbf{bold}. ME latency includes model editing and reasoning, while ICE latency includes knowledge retrieval and reasoning.}
\vspace{-8pt}
\label{tab:latency}
\arrayrulecolor{black}
\end{table}

\section{Related Work}

\paragraph{LLMs' Hallucination}

Pre-training on large-scale corpora equips LLMs with extensive parametric memory, including commonsense and factual knowledge ~\citep{petroni2019languagemodelsknowledgebases, li2022systematicinvestigationcommonsenseknowledge}. However, this parametric knowledge may be inaccurate due to errors or outdated information in the pre-training data, leading to hallucinations~\citep{zhang2023sirens, huang2023survey, wang2023survey} where the content generated by LLMs deviates from established world knowledge.

\paragraph{Knowledge Conflict}

To mitigate the hallucinations, tools~\citep{nakano2022webgptbrowserassistedquestionansweringhuman, yao2023reactsynergizingreasoningacting, qin2024toollearningfoundationmodels} or retrieval-augmented methods~\citep{guu2020realmretrievalaugmentedlanguagemodel, izacard2021leveragingpassageretrievalgenerative, zhong2022traininglanguagemodelsmemory}, such as ChatGPT Plugins and New Bing, have been proposed as effective solutions to provide external knowledge evidence.  However, external knowledge may inevitably conflict~\citep{petroni2020contextaffectslanguagemodels, si2023promptinggpt3reliable, xie2024adaptivechameleonstubbornsloth} with parametric knowledge, leading to unreliable support, especially when LLMs are overly confident in their own parametric knowledge.

\paragraph{Knowledge Editing}

KE \citep{yao2023editing} has been proposed to update outdated information, enabling models to answer current questions accurately. In general, existed KE can be divided into two main categories. 
ME \citep{zhu2020modifying, meng2022locating, meng2022mass, huang2023transformer} involves modifying model parameters or structure to prevent undesired outputs. 
ICE \citep{mitchell2022memory, madaan2022memory, zhong2023mquake, zheng2023can} edit knowledge by prompting LLMs with the newly updated facts.
However, both approaches are affected by localization or knowledge conflicts, leading to hallucinations. 

\section{Conclusion}
In this paper, we proposed a new improved baseline for knowledge editing, called \ours. 
Unlike the locate-and-edit KE approaches such as ME and ICE, \ours removes all parametric knowledge, regardless of whether it conflicts with new knowledge.
By leveraging LLMs to extract entities and relations from the original question, \ours performs multi-hop reasoning over up-to-date knowledge structures to derive accurate answers. 
This new paradigm offers higher editing accuracy, faster performance, and greater robustness.
Our work paves the way for further advancements in knowledge editing.

\section*{Limitations}
This work presents an improved baseline for KE.  Unlike ME and ICE, which rely on text-based editing information, \ours requires a more structured knowledge format to support LLM reasoning.  The decline from the editing tasks in Table \ref{tab:observation} to the QA tasks in Tables \ref{tab:open} and \ref{tab:close} reflects the loss in entity and relation extraction for \ours.  
While \ours demonstrates strong robustness, there is still a noticeable decline as reasoning hops and the number of edited instances increase, indicating potential errors in LLM reasoning.  

\section*{Ethical Considerations}
In this study, we adhere to ethical guidelines by using only open-source datasets and employing models that are either open-source or well-established in the scientific community. We utilize counterfactual public datasets for knowledge editing to evaluate knowledge updates. Our proposed \ours method focuses on updating knowledge to enable LLMs to accurately answer real-world questions. We are committed to maintaining high ethical standards throughout our research, emphasizing transparency and promoting the responsible use of technology for the betterment of society.

\bibliography{custom}

\appendix

\end{document}